\documentclass[conference]{IEEEtran}
\usepackage{times}

\usepackage[numbers]{natbib}
\usepackage{multicol}
\usepackage[bookmarks=true]{hyperref}
\usepackage{ulem}
\usepackage{amsmath}
\usepackage{amssymb}
\usepackage{color}
\usepackage{graphicx} 
\usepackage{algorithm}
\usepackage{algpseudocode}
\usepackage[utf8]{inputenc} 
\usepackage[T1]{fontenc}    
\usepackage{ulem}
\usepackage{hyperref}       
\usepackage{url}            
\usepackage{booktabs}       
\usepackage{amsfonts}       
\usepackage{amsmath}       
\usepackage{nicefrac}       
\usepackage{microtype}      
\usepackage{xcolor}         
\usepackage{graphicx}
\usepackage{subfigure}
\usepackage{booktabs}
\usepackage{multirow}
\usepackage{graphicx} 
\usepackage{graphicx}
\usepackage{booktabs} 
\usepackage{subcaption}
\usepackage{caption}

\newcommand{\R}{{\mathbb R}}
\def\vec#1{\mathbf{#1}}
\def\dt#1{\dot{#1}}

\begin{document}

\title{Neural Lyapunov and Optimal Control}

\author{Daniel Layueghi, Steve Tonneau, Michael Mistry}

\author{\authorblockN{Daniel Layeghi}
\authorblockA{
School of Informatics\\
The University of Edinburgh\\}
\and
\authorblockN{Steve Tonneau}
\authorblockA{
School of Informatics\\
The University of Edinburgh\\}
\and
\authorblockN{Michael Mistry}
\authorblockA{
School of Informatics\\
The University of Edinburgh\\}}


%

\maketitle

\begin{abstract}

Despite impressive results, reinforcement learning (RL) suffers from slow convergence and requires a large variety of tuning strategies.
In this paper, we investigate the ability of RL algorithms on simple continuous control tasks. We show that without reward and environment tuning, RL suffers from poor convergence. In turn, we introduce an optimal control (OC) theoretic learning-based method that can solve the same problems robustly with simple parsimonious costs. We use the Hamilton-Jacobi-Bellman (HJB) and first-order gradients to learn optimal time-varying value functions and therefore, policies. We show the relaxation of our objective results in time-varying Lyapunov functions, further verifying our approach by providing guarantees over a compact set of initial conditions. We compare our method to Soft Actor Critic (SAC) and Proximal Policy Optimisation (PPO). In this comparison, we solve all tasks, we never underperform in task cost and we show that at the point of our convergence, we outperform SAC and PPO in the best case by 4 and 2 orders of magnitude.  
\end{abstract}
\IEEEpeerreviewmaketitle

\section{Introduction}
Finding the optimal law to control a non-linear dynamical system with respect to an objective function is an open challenge for many systems. 
In recent years a class of reinforcement learning (RL) algorithms have empirically demonstrated an ability to to solve complex continuous control tasks \cite{hwangbo2019learning, andrychowicz2020learning}. RL does so by leveraging concepts within optimal control (OC) by parameterising and learning the policy or the value function space. Many of these impressive feats however, rely on strategies to improve convergence and stability such as; linearising dynamics regimes via Proportional-Derivative (PD) control or avoiding non-linear rspacesegimes by terminating the episode outside of locally linear state space\cite{8793865}\cite{tassa2012synthesis}. Additionally, they rely on complex reward-shaping and extensive hyperparameter tuning\cite{henderson2018deep}. Part of the reason RL requires these strategies is that many of the algorithms use zeroth-order gradient estimates to minimise objectives. These estimates are high in variance and can descend into regions of the state space that impair convergence \cite{suh2022differentiable}.

Trajectory optimisation (TO) is another class of optimal control that has proven to be effective and efficient in problems where dynamics are nonlinear and unstable\cite{tassa2012synthesis, kuindersma2016optimization}. The efficiency of TO methods comes from leveraging first or higher-order derivatives of the objective. This results in more stable optimisation that does not require extensive hyperparameters, complex cost design or termination strategies. However, TO methods parametrise trajectory spaces and require reoptimisation per new initial condition. 

\begin{figure}[t]
  \centering
  \includegraphics[width=1\linewidth]{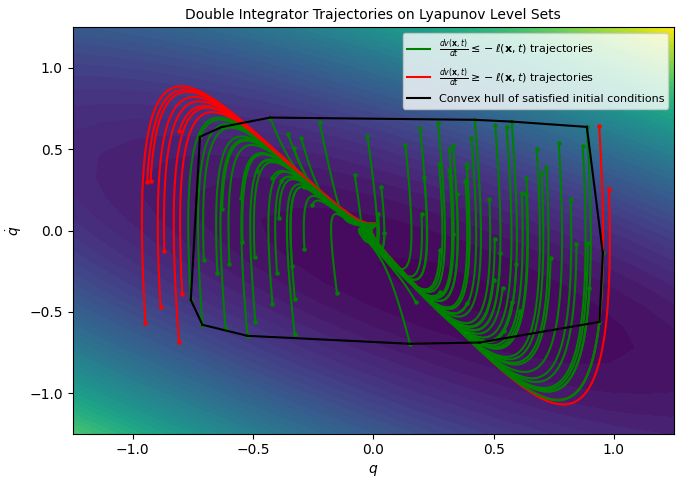}
  \caption{Compact stability region for double integrator, computed by Neural Lyapunov Control.}
  \label{fig:DI Lyap}
\end{figure}

In this paper, we investigate the possibility of leveraging OC-theoretic tools to learn an optimal policy that allows us to not require online reoptimisation while avoiding the drawbacks of reinforcement learning. This requires finding a way to leverage Optimal Control theory within massively parallel machine learning frameworks. We ask two questions:

\begin{itemize}
\item Can RL algorithms solve simple continuous control tasks without the need for dampening the dynamics, shaping rewards or avoiding regions of the state space?
\item Can we solve these problems with minimal hyperparameter optimisation and reward shaping while embracing the entirety of the state and control space?
\end{itemize}

We believe the above are important problems to consider to improve on the robustness, applicability and reproducibility of learning-based approaches.
To this end, we formulate a new OC theoretic function parameterised approach to learn optimal control policies. We utilise differentiable dynamics and the Hamilton-Jacobi-Bellman (HJB) optimality constraint to formulate mathematical programs that enable learning value functions and therefore, optimal policies. Additionally, we show that a specific relaxation of this objective allows us to learn Lyapunov functions, enabling us to further verify the stability of our method over a compact set of initial conditions. 

To solve these programs, we leverage neural ODEs \cite{sandoval2022neural}, a new gradient estimator, and the parallel optimisation capabilities of deep learning. Finally, we compare our method to Soft Actor-Critic (SAC)\cite{haarnoja2018soft} and Proximal Policy Optimisation (PPO)\cite{schulman2017proximal} on selected linear and nonlinear control affine tasks. We employ minimal cost shaping by using simple parsimonious quadratic costs. We do not restrict the landscape of the state space by early termination. Additionally, we use identical network architectures, episode horizons and boundaries of initial conditions. 
We empirically show the following:
\begin{itemize}
    \item RL suffers from poor convergence on environments where minimal reward shaping and environment tuning are used.
    \item In our experiments our proposed method solves the tasks with significantly faster convergence and variance in random seeds. Outperforming SAC and PPO by at least a factor of 74 and 2 respectively.
\end{itemize}
 
\section{Related Work}
\subsubsection{RL robustness}\label{tricks}
Despite RL's effectiveness, several works have studied the shortcomings of the popular approaches. Authors in \cite{henderson2018deep} performed a comprehensive study on the reproducibility of policy gradient algorithms such as PPO, TRPO and DDPG \cite{schulman2017proximal, schulman2015trust, lillicrap2015continuous}. Their results showed that these algorithms are very sensitive to a variety of parameters. For example, the choice of random seed may lead to significant outperformance on the same solver. The reward is also crucial to the performance, and authors show simple scaling of the rewards may lead to policies that fail entirely. The choice of network architecture and activation function have also been shown to be consequential. The sample complexity of RL based methods has also been in studied. In \cite{recht2019tour} authors show that policy gradient methods require a high number of samples to converge even on simple linear LQR problems. They also show that applying LQR to simple learnt models outperform policy gradient based methods by orders of magnitude. The source of this sample complexity is also investigated by \cite{suh2022differentiable}. Authors show zeroth-order gradient estimates, the underlying estimator in policy gradient methods, are very high variance and inefficient for a variety of continuous control tasks. Thus this raises an important drawback of RL, namely its sample complexity and sensitivity to hyper parameters. As a result, many approaches have been proposed that aim to mitigate these problems. 
\subsubsection{OC theoretic policy/value learning}
Approaches to learning OC policy and value functions have been previously explored. Many seminal works exist in offline settings, where off-the-shelf solvers are used to solve OC problems, and the generated data is then used for training. For example \cite{mordatch2015interactive} applied this formulation to character control. The authors used second-order nonlinear OC to generate state trajectories. A neural network was then used to train a policy that can generate these trajectories and perform complex dynamic behaviour. However, the implicit consequential constraints within the trajectory optimisation problem, such as inverse dynamics, are not considered within the regression. Additionally, this process is computationally expensive and requires compute clusters for solving multiple OC problems in parallel. 
Other works within the offline setting focus on the value function space. \cite{9981234} formulate a supervised approach that uses neural networks to learn to fit approximate cost-to-gos, collected from trajectory optimisation. The authors tackle the complex task of humanoid foothold selection. This approach, inspired by \cite{9361258}, is based on the notion that informed terminal constraints and value functions can reduce planning horizons. However, due to the task complexity and the difficulty of the data generation process, the authors only consider two initial conditions. Additionally, as they solve a supervised problem, implicit constraints such as Bellman backup are not typically considered in the regression. 

Work done by \cite{lutter2021value} addresses the problem of implicit constraints. They do so by approaching the problem from a value iteration perspective. In this case, the one-step HJB policy is used to rollout and compute cost-to-go data that is then fitted to the value function using Bellman backup loss. Authors of \cite{lowrey2018plan} also employ a very similar approach, however, the HJB policy is replaced with a trajectory optimiser. Both authors however, do not consider time parameterisation of the value function, even though the data is collected over a finite time.  Overall, offline approaches are computationally intensive and primarily rely on supervised training methods that overlook the implicit dynamics of data being learnt.

Other online and more unified approaches have formulated methods that learn Lyapunov and control barrier functions. These methods train neural networks to inherently satisfy Lyapunov stability by minimising the penalty form of Lyapunov constraints. \cite{dawson2022safe} jointly optimises over Lyapunov and a policy network, whilst \cite{10077790} additionally optimises over a barrier function. Authors in \cite{chang2019neural} also use a similar approach whilst introducing a verification procedure for certifying the learnt functions. However, these methods do not apply to finite time horizon problems such as trajectory optimisation, as they do not solve for time-varying Lyapunov functions. Similarly the finite horizon formulation of this approach is also investigated in \cite{2021arXiv211111277X}. However, similar to before this approach requires offline data generation by nominal controllers.
Perhaps the closest line of work is done by \cite{ainsworth2021faster} where the authors exploit the structure of neural ODEs with application to policy learning, given known dynamics. Similarly to \cite{ainsworth2021faster}, \cite{sandoval2022neural} explores the constrained version of the policy-based formulation. \cite{zhang2023initial} considers the time parameterised version of the policy, whilst introducing further regularisation using offline data. However, the above methods are formulated in policy space and cannot be applied to value and Lyapunov functions synthesis. 

More specific HJB based approaches also exist. Authors in \cite{bansal2021deepreach} focuses on reachability analysis, wherein they relax the Hamilton-Jacobi (HJ) differential equation and employ neural networks to approximate the corresponding value function, effectively mitigating the curse of dimensionality typically associated with classical grid-based solvers. The results however are not compared to any baselines and it is unclear whether backpropogation through time is feasible for high dimensional systems. In a similar vein, \cite{engin2023neural} utilises a composite loss involving the HJB equation, Hamiltonian, and trajectory cost. This loss is minimised by learning both value and policy functions. The efficacy of this method is demonstrated through comparisons with model-based RL. To best of our understanding this method both learns a policy and a value function independently so the effect of each network is unclear. In \cite{Zhong2020Symplectic} authors also consider a similar domain where they learn the costate vector in the maximum principle formulation. However, this vector only considers the value function in its time differential space making the learnt function limited to the trajectories on which it was trained.

In this work, we introduce a unified approach that solves finite-time parameterised OC problems while leveraging neural ODEs to learn the corresponding time-varying Lyapunov and value functions. We test our method on linear and nonlinear control-affine systems. We compare to RL baselines outperforming all by orders of magnitude.
\section{Preliminaries}
\subsection{Optimal Control: } Optimal control is grounded within the Hamilton-Jacobi-Bellman (HJB) equation.
\begin{equation}
-v_t\left(\vec{x}\left(t\right), t\right) = \ell\left(\vec{x}\left(t\right), \vec{u}\left(t\right)\right) + \vec{f}\left(\vec{x}\left(t\right), \vec{u}\left(t\right)\right)^\top v_{\vec{x}}\left(\vec{x}\left(t\right), t\right)
\end{equation}
where $\dt{\vec x} = \vec f(\vec x, \vec u)$ represents the deterministic dynamics, given state $\vec x \in \R^n$ and control $\vec u \in \R^m$. $\ell(\vec x, \vec u)$ represents the state-control cost, $v_t(\vec x(t), t)$ and $v_\vec x(\vec x(t), t)$ represent the Legendre notation for partial derivatives of the value function with respect to time and state, respectively. This is a Partial Differential Equation (PDE) defining the time evolution of the value function. In other words the compact description of the relationship between optimal costs for different states and times. In the case of the existence of a $C^1$ value function, cost and affine dynamics, one can analytically compute a \textit{closed loop} feedback law optimal for all initial conditions $\vec {x}(0)$. The following shows this optimal control for a control-affine dynamical system where $\vec{\dot{x}} = \vec f(\vec x(t)) + \vec g(\vec x(t))\vec u(t)$. Given a quadratic regularisation of control $\|\vec u\|_{\vec R}$. Where $\mathbf{R}$ is a positive definite matrix.
\begin{equation}
\begin{aligned}
\label{eq: hjb policy}
\operatorname*{min}_{\vec u \in \vec U} & \left [v_t(\vec x(t), t) + \ell(\vec x(t), \vec u(t)) + \vec f(\vec x(t), \vec u(t))^\top v_{\vec x}(\vec x(t), t)\right ] \\
& \pi ^* (\vec x(t), t) = -\frac{1}{2} \vec {R}^{-1}\vec{g}(\vec x(t))^\top v_{\vec x}(\vec x(t), t)
\end{aligned}
\end{equation}
The above policy gives us a time-varying optimal policy $\vec{\pi^*}(\vec x(t), t)$ for any initial condition. This is an important property of the HJB equation which relies on the strong assumption of the existence of a known value function.
On the other hand, Pontryagin Minimum Principle (PMP) alleviates this problem by operating in the tangential space. It converts this PDE into an Ordinary Differential Equation (ODE) by interpreting $v_\vec x$ the gradient of the value as a stand-alone vector $\vec{p}$ known as a costate vector. Referring to the right-hand side of equation \ref{eq: hjb policy} as Hamiltonian $H$, the PMP aims to minimise $H$ over a horizon $T$ using an optimal control trajectory $\vec{u^*}$. This trajectory is determined given an initial condition $\vec{x}(0)$ and a terminal condition $\ell_{\vec{x}}(\vec{x}_T) = \vec{p}(T)$, subject to the following:
\begin{equation}
\begin{aligned}
\vec {\dot{x}} &= H_{\vec p}|_*, \hspace{2em} \vec{\dot{p}^*} = -H_{\vec x}|_*  \\
\vec{u^*} (t) &= \operatorname*{arg\,max}_{\vec u \in \vec U} H(\vec{x^*}(t), \vec{u}, \vec{p^*}(t))
\end{aligned}
\end{equation}
Due to its tractability, this framework underpins the majority of trajectory optimisation algorithms such as iterative LQR \cite{tassa2012synthesis} or differential dynamic programming (DDP) \cite{jacobson1968new}. However, it only generates an open loop trajectory valid for a single initial condition. This can be a significant limitation as new initial conditions require reoptimisation. For further discussion around these topics, we refer the reader to \cite{liberzon2011calculus}.
\subsection{Neural ODEs} Neural ODEs are both practically and theoretically\cite{chen2018neural} fundamental to our work. In this section, we briefly describe neural ODEs from an OC perspective. The general objective for neural ODEs is defined as:
\begin{equation}
L\left(\vec x(T)\right) = L \left(\vec x(0) + \int_{t_0}^{T} \vec f\left(\vec x(t), \vec u, t\right) dt\right)
\end{equation}
Neural ODEs differ from traditional OC methods in two ways. Firstly. the control parameter $\vec u$ is a time-invariant variable interpreted as network weights. Secondly, the loss is evaluated at the end of the trajectory due to the analogous nature of the integration time step to hidden layers in a standard neural network. As a result, the canonical equations of neural ODEs are a special case of PMP, with the costate vector $\vec{\dot{p}^*} = -H_{\vec x}|_* = -\vec f_{\vec x}^\top(\vec x(t), \vec \theta, t)\vec p(t)$ ignoring the effects of running loss $L_{\vec x}(\vec x(t))$. Therefore, although neural ODEs create a familiar grounding to OC they cannot be immediately applied to problems where a change of state over time matters. The authors of \cite{ainsworth2021faster} provide a reformulation of neural ODE's gradient estimator for policy learning, which we show can be extended to the value and Lyapunov space in the next section.
\section{Learning Lyapunov and value functions}\label{method}
Our approach provides a straightforward and effective framework for learning value and Lyapunov parameterised OC problems while respecting implicit OC constraints.
\subsection{Value functions}
Let us focus on the HJB equation \ref{eq: hjb policy} under the optimal policy $\pi ^* (\vec x(t), t)$. By moving the inner product between $v_{\vec x}$ and the dynamics $\vec f(\vec x(t), \vec u(t))$ or $\frac{d\vec x(t)}{dt}$ to the left-hand side we can rewrite HJB as a definition for the total rate of change of the value function
\begin{equation}
\begin{aligned}
\label{eq: dvdt}
\frac{\partial v(\vec x(t), t)}{\partial t} + \frac{\partial v(\vec x(t), t)}{\partial \vec x(t)}\frac{d \vec x(t)}{dt} &= -\ell(\vec x(t), \pi ^* (\vec x(t))) \\
\frac{dv(\vec x(t), t)}{dt} &= -\ell(\vec x(t), \pi ^* (\vec x(t)))
\end{aligned}
\end{equation}
The above can be interpreted as a constraint on the rate of change of the value function under the optimal policy. Integrating over a horizon T allows us to evaluate the consistency of optimal policy with respect to the above constraint. This leads to:
\begin{equation}
\begin{aligned}
\label{eq: int dvdt}
\int_0^T\frac{dv(\vec x(t), t)}{dt} &= -\int_0^T \ell(\vec x(t), \pi ^* (\vec x(t))) \\
v(\vec x(T), T) - v(\vec x(0), 0) &= -\int_0^T \ell(\vec x(t), \pi ^* (\vec x(t)))
\end{aligned}
\end{equation}
Equation \ref{eq: int dvdt}, provides us with an equality that defines the evolution of the value function over the horizon $T$ as a function of the running loss. However, we aim to learn the granular temporal and spatial change in the value function over the full horizon. In order to capture this effect we discretise both integrals over $\Delta t$ increments and apply the left Riemann sum approximation. 
\begin{equation}
\begin{aligned}
\label{eq: disc int dvdt}
\int_t^{t+\Delta t}\frac{dv(\vec x(t), t)}{dt} &= -\int_t^{t+\Delta t} \ell(\vec x(t), \pi ^* (\vec x(t), t)) \\
v(\vec x_{t+\Delta t}, t+\Delta t) - v(\vec x_t, t) &\simeq -\Delta t \times \ell(\vec x_t, \pi ^* (\vec x_t, t))
\end{aligned}
\end{equation}
equation \ref{eq: disc int dvdt} provides an approximate temporal and spatial instantaneous constraint on the value function. By rearranging and squaring we can convert this constraint into a soft penalty $P_v$ where:
\begin{equation}
\begin{aligned}
\label{eq: value pen}
P_v = \bigl( & v\left(\vec x_{t+\Delta t}, t+\Delta t\right) - v\left(\vec x_t, t\right) \\
&+ \Delta t \times \ell \left(\vec x_t, \pi ^* \left(\vec x_t, t\right)\right) \bigr)^2
\end{aligned}
\end{equation}
The mathematical program \ref{eq: value prog} leverages this penalty to learn an approximate value function $\Tilde{v}(\vec x, t; \theta)$ that minimises the integration of this penalty over the horizon $T$ and a compact set of initial conditions $\vec x_0 \in X_0$ where $|X_0| = K$. Program \ref{eq: value prog} uses the optimal feedback policy shown in \ref{eq: hjb policy} and is therefore valid only for control-affine dynamics and quadratic regularisation of controls. However analytical optimal policies may be computed under any convex and PSD control constraints and are therefore easily incorporated within this framework. This is further discussed in Section~\ref{sec:disc}.
\begin{equation}
\begin{aligned}
\label{eq: value prog}
&\min_{\theta}\biggl[\frac{1}{K}\sum_{k = 0}^K\sum_{n = 0}^{N-1}\Bigl(\Tilde{v}\left(\vec{x}_{n+1, k}, n+1; \theta\right) \\
&\qquad - \Tilde{v}\left(\vec{x}_{n, k}, n; \theta\right) + \Delta t \times \bigl(\|\vec u^*_{n, k}\|_\vec R + \ell\left(\vec{x}_{n, k}\right)\bigr)\Bigr)^2\biggr] \\ 
&\text{s.t:}\\
&\quad \vec{\dot{x}}(t) = \vec f(\vec x(t)) + \vec g(\vec x(t))\vec u^*(\vec x, t) \\
&\quad \vec u^*(\vec x(t), t) = -\frac{1}{2} \vec {R}^{-1}\vec{g}(\vec x(t))^\top v_{\vec x(t)}(\vec x(t), t; \theta)
\end{aligned}
\end{equation}
The above mathematical program defines a finite-time parameterised OC problem that aims to learn a value function that approximately minimises the Bellman backup condition in equation \ref{eq: disc int dvdt} over horizon $T$, where \(N = \frac{T-t_0}{\Delta t}\) represents the number of discrete time steps in the interval from \(t_0\) to \(T\). Each \(t_n = n\Delta t + t_0\) and \(t_{n+1} = (n+1)\Delta t + t_0\) specify the time instances at the \(n\)-th and \(n+1\)-th time steps, respectively. Here, \(\vec x_{n, k}\) and \(\vec u^*_{n, k}\) denote the state and control input at time \(t_n\).
\subsection{Lyapunov functions}
We briefly motivate constructing Lyapunov programs. In program \ref{eq: value prog} our objective is to minimise the square Bellman error. While it is possible to minimise, approximation errors may lead to unquantifiable loss of performance in trajectory cost. The Lyapunov function relaxes this constraint to an inequality, thereby, allowing us to convert the problem to a satisfability condition. As a result, we are able to provide performance guarantees on compact regions of the state space, which are not possible for program \ref{eq: value prog}.

We now focus on the derivation of the Lyapunov program. Finite-time Lyapunov analysis states that if we have a continuously differentiable positive definite function $v(\vec x(t), t)$, where $v(\vec x(t), t) > 0$ for $\vec x(t) \neq 0$ and $v(0, t) = 0$ then $\vec f(\vec x(t), t)$ is stable if:
$\dot{v}(\vec x(t), t) = \frac{\partial v(\vec x(t), t)}{\partial \vec x(t)} \vec f(\vec x) + \frac{\partial v(\vec x(t), t)}{\partial t} < 0, \
\dot{v}(0, t) = 0$. This condition must hold for all $\vec x$ and all $t$. However, finding such Lyapunov functions is not trivial and is unknown apriori. Equation \ref{eq: dvdt} and the conditions above impose constraints on the rate of change of function $v$ in both the contexts of value and Lyapunov functions. However, the Lyapunov condition is a relaxed version of the HJB equality constraint. Constraint \ref{eq: dvdt} is a lower bound where satisfying it transforms $v$ from a Lyapunov function into an optimal value function. We aim to learn an approximate Lyapunov function that allows for generating controls that can sub-optimally complete a task but also satisfy Lyapunov constraints within a compact set of initial conditions. To formulate this we relax the HJB condition and define the Lyapunov constraint with respect to a task loss $\ell(\vec x)$.
\begin{equation}
\begin{aligned}
\label{eq: dvdt<}
\frac{\partial v(\vec x(t), t)}{\partial t} + \frac{\partial v(\vec x(t), t)}{\partial \vec x(t)}\vec f(\vec x(t), \vec u(t)) &\leq -\ell(\vec x(t), \vec u(t)) \\
\frac{dv(\vec x(t), t)}{dt} &\leq -\ell(\vec x(t), \vec u(t))
\end{aligned}
\end{equation}
Performing the same integration over discrete time step $\Delta t$ we obtain an instantaneous inequality constraint that can be converted to a soft penalty $P_L$:
\begin{equation}
\begin{aligned}
\label{eq: lyap pen}
P_L = \max\bigl( & v\left(\vec x_{t+\Delta t}, t+\Delta t\right) - v\left(\vec x_t, t\right) \\
&+ \Delta t \times \ell \left(\vec x_t, \vec u \left(\vec x_t, t\right)\right), 0 \bigr)
\end{aligned}
\end{equation}
The program \ref{eq: lyap prog} uses this penalty to formulate an objective function that aims to learn an approximate Control Lyapunov Function (CLF) that enforces this Lyapunov constraint over a horizon $T$ and $K$ set of initial conditions.
\begin{equation}
\begin{aligned}
\label{eq: lyap prog}
&\min_{\theta}\biggl[\frac{1}{K}\sum_{k = 0}^K\sum_{n = 0}^{N-1}\max\Bigl(\Tilde{v}\left(\vec{x}_{n+1, k}, n+1; \theta\right) \\
&\qquad - \Tilde{v}\left(\vec{x}_{n, k}, n; \theta\right) + \Delta t \times \bigl(\|\vec u^*_{n, k}\|_\vec R + \ell\left(\vec{x}_{n, k}\right)\bigr), 0\Bigr)\biggr] \\ 
&\text{s.t:}\\
&\quad \vec{\dot{x}}(t) = \vec f(\vec x(t)) + \vec g(\vec x(t))\vec u^*(\vec x, t) \\
&\quad \vec u^*(\vec x, t) = -\frac{1}{2} \vec {R}^{-1}\vec{g}(\vec x)^\top v_{\vec x}(\vec x, t; \theta)
\end{aligned}
\end{equation}
We make use of the same HJB feedback policy used in value learning. This is because, as we show in the Appendix, the policy is a valid CLF for a given Lyapunov function under the same affinity and regularisation assumptions as program \ref{eq: value prog}. We show this in Theorem 1 of the supplementary material. However, other choices for a CLF such as Sontag's universal controller \cite{lin1991universal} are also valid. Subscripts $n$ and $k$ represent the same parameters as program \ref{eq: value prog}. To parameterise the Lyapunov function while respecting Lyapunov constraints we make use of input convex neural networks (ICNNs) \cite{amos2017icnn} with an additional positive constant term to approximate the positive definite Lyapunov function: $v(\vec x, t; \theta) = v_\text{ICNN}(\vec x, t; \theta) + \epsilon \|\vec x \|_2^2$.
\subsubsection{Neural ODE}
The mathematical programs \ref{eq: value prog} and \ref{eq: lyap prog} have a bounded time horizon. While the neural ODE gradient estimator are suitable for finite horizon optimisation, they cannot be directly applied to these programs as the task loss needs to be evaluated over every timestep of the entire horizon T. To alleviate this, we follow the same modifications shown in \cite{ainsworth2021faster}. These modifications aim to convert the canonical equations of neural ODEs to the PMP case.
\begin{equation}
\begin{aligned}
\label{eq: node pmp}
\dot{\vec a} &= - \vec{a}(t)^\top \vec f_{\vec x(t)}(\vec{x}, \vec{u}(\vec x(t), t, \theta)) \\
&\quad - \ell_{\vec x}(\vec x(t), \vec{u}(\vec x(t), t; \theta)) \\
\dot{\ell}_{\theta} &= - \vec{a}(t)^\top \vec f_{\vec u}(\vec{x}(t), \vec{u}(\vec x(t), t, \theta))\vec{u}_{\theta}(\vec x(t), t, \theta) \\
&\quad - \ell_{\vec u}(\vec{x}(t), \vec{u}(\vec x(t), t; \theta))\vec{u}_{\theta}(\vec x(t), t, \theta)
\end{aligned}
\end{equation}
These canonical equations differ from vanilla neural ODEs mentioned in \cite{chen2018neural}. The resulting gradient estimator is less efficient in complexity compared to its vanilla form. However, it is a necessary addition.

We have shown how parameterised OC problems can be formulated using the mathematical programs \ref{eq: lyap prog} and \ref{eq: value prog}. In the next segment, we will demonstrate how these problems can be solved using the gradient estimator \ref{eq: node pmp}. We obtain these results on systems with control-affine dynamics.
\section{Empirical results}
\subsubsection{Criteria}
\begin{figure*}[t]
  \captionsetup{skip=5pt, belowskip=-5pt} 
  \centering
  \begin{minipage}{0.23\textwidth}
    \includegraphics[width=\textwidth]{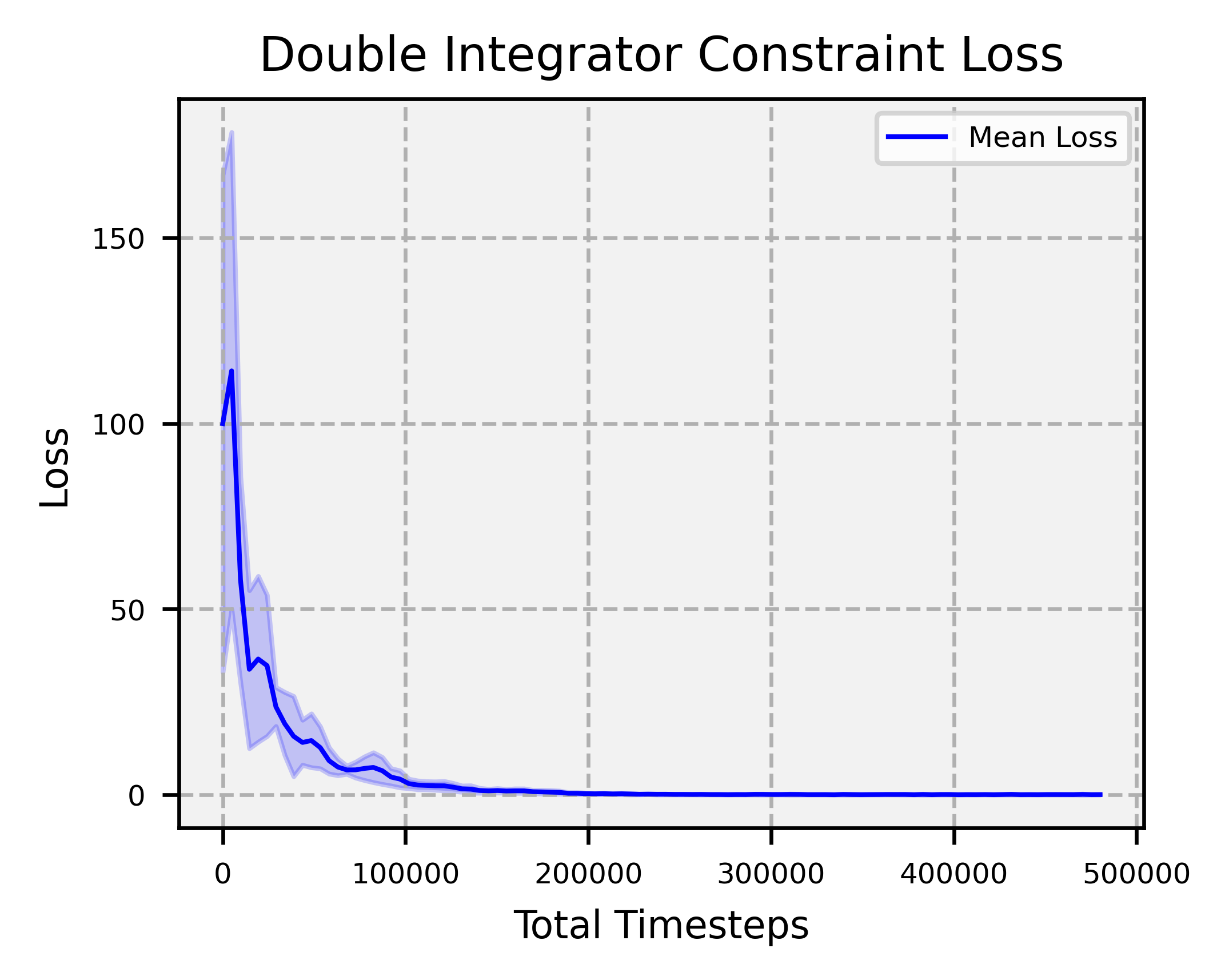}
  \end{minipage}\hfill
  \begin{minipage}{0.23\textwidth}
    \includegraphics[width=\textwidth]{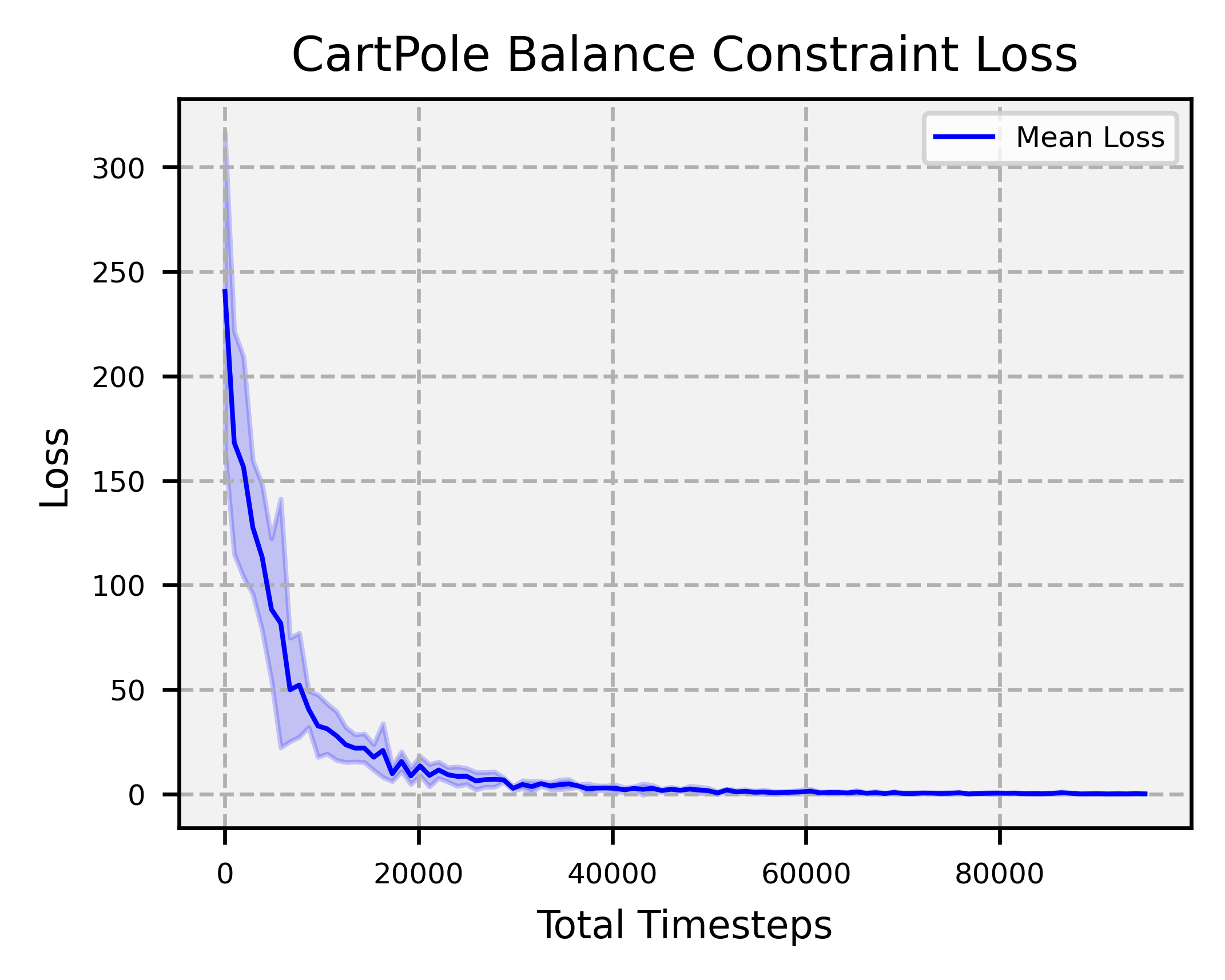}
  \end{minipage}\hfill
  \begin{minipage}{0.23\textwidth}
    \includegraphics[width=\textwidth]{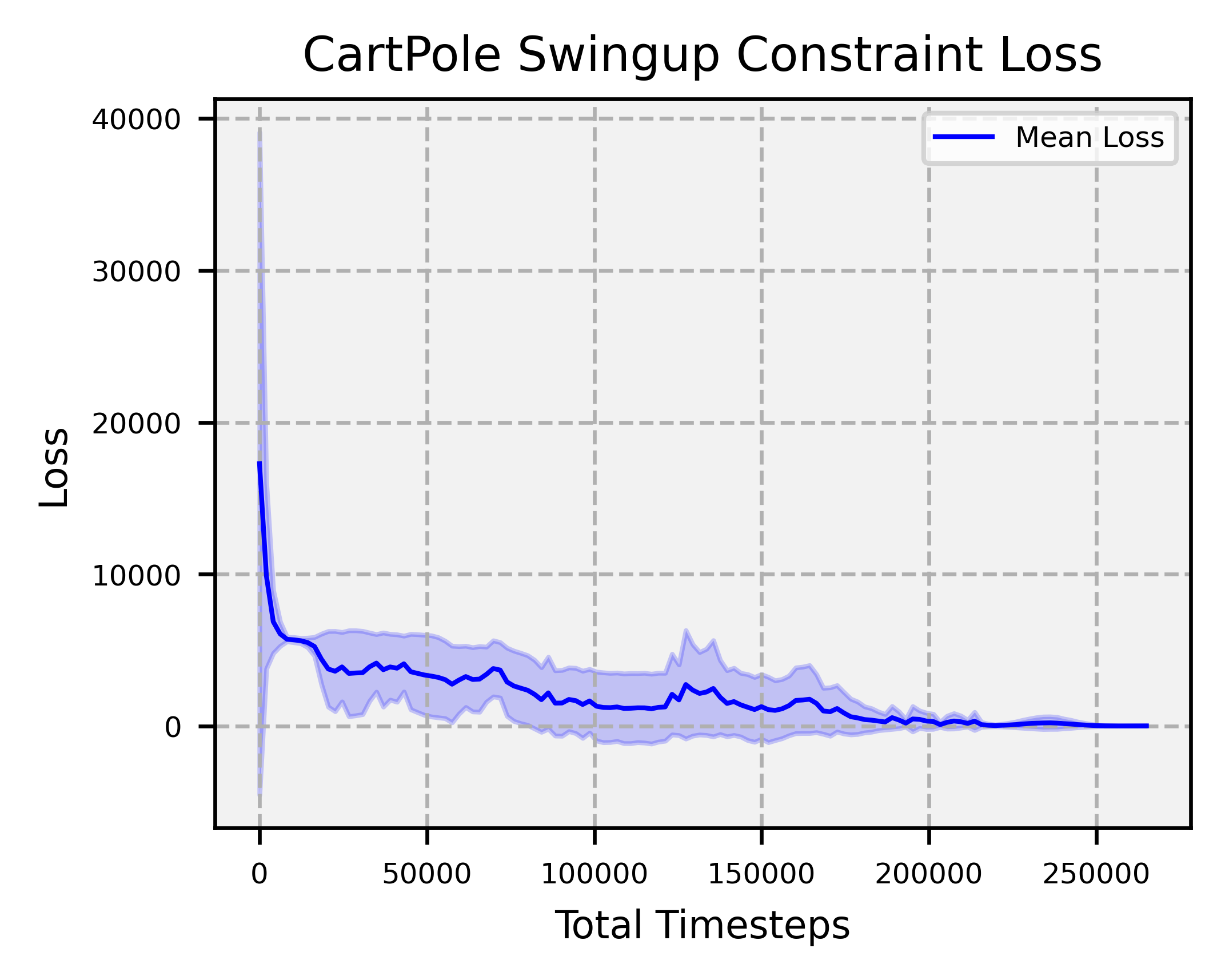}
  \end{minipage}\hfill
  \begin{minipage}{0.23\textwidth}
    \includegraphics[width=\textwidth]{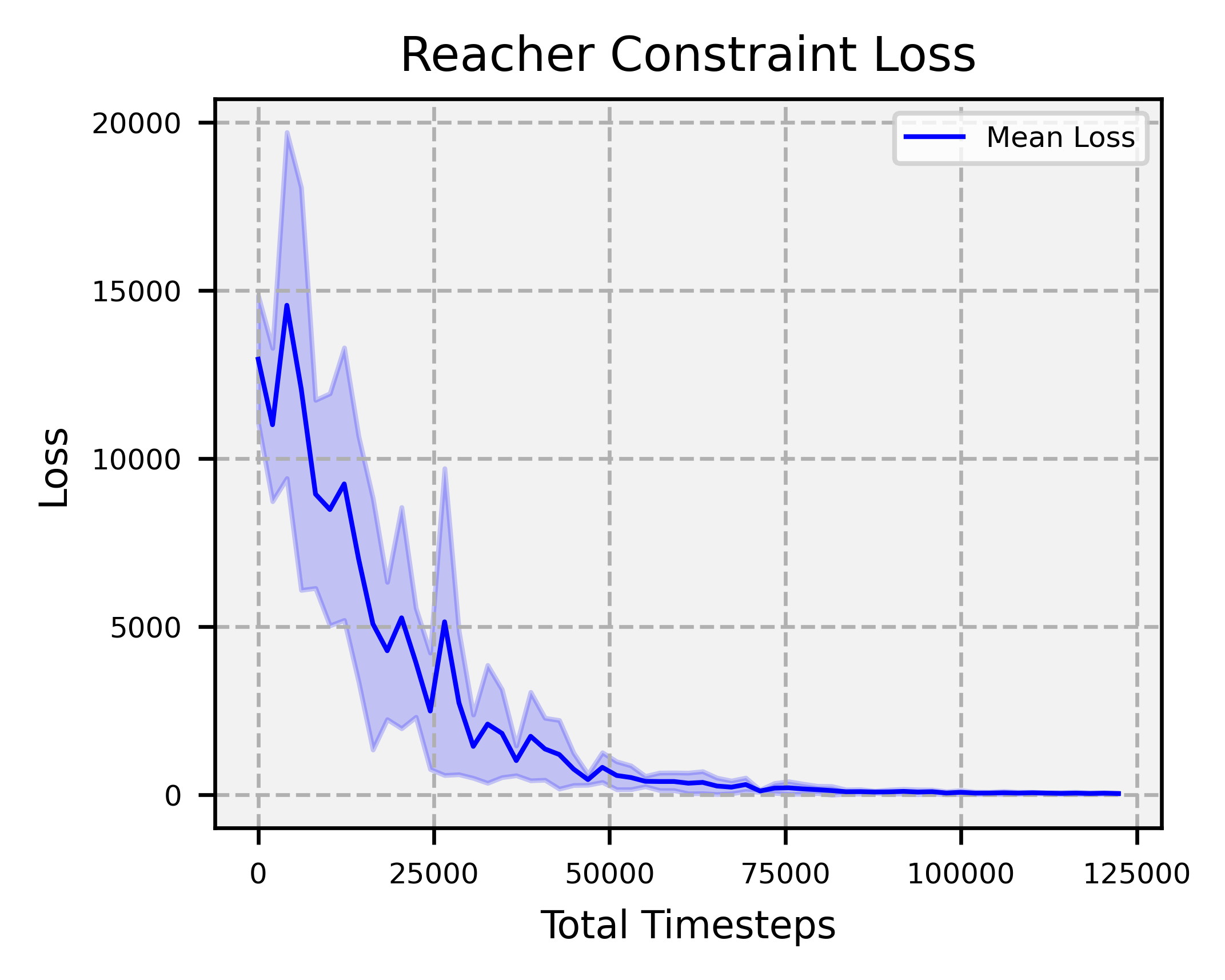}
  \end{minipage}
  
  \begin{minipage}{0.23\textwidth}
    \includegraphics[width=\textwidth]{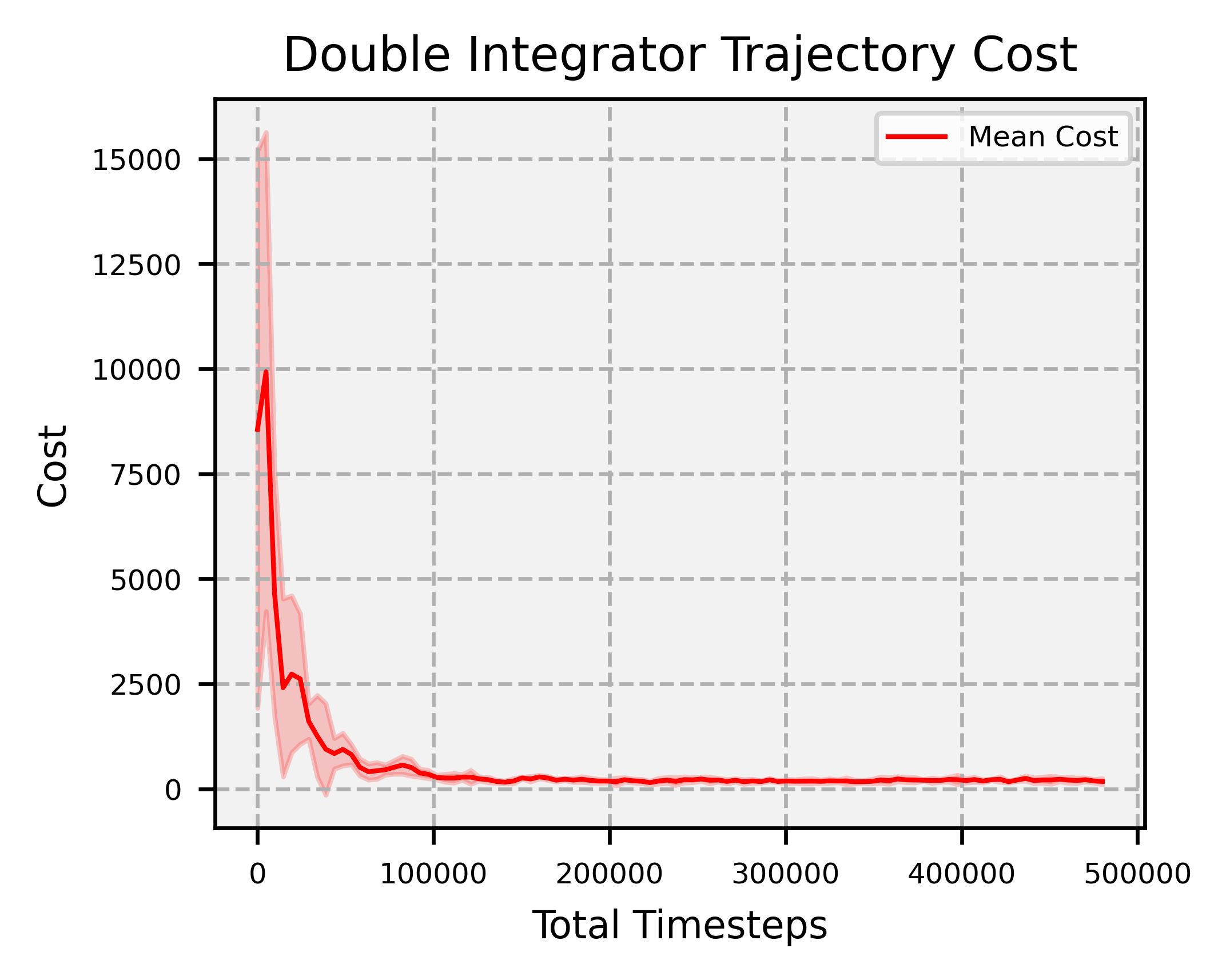}
  \end{minipage}\hfill
  \begin{minipage}{0.23\textwidth}
    \includegraphics[width=\textwidth]{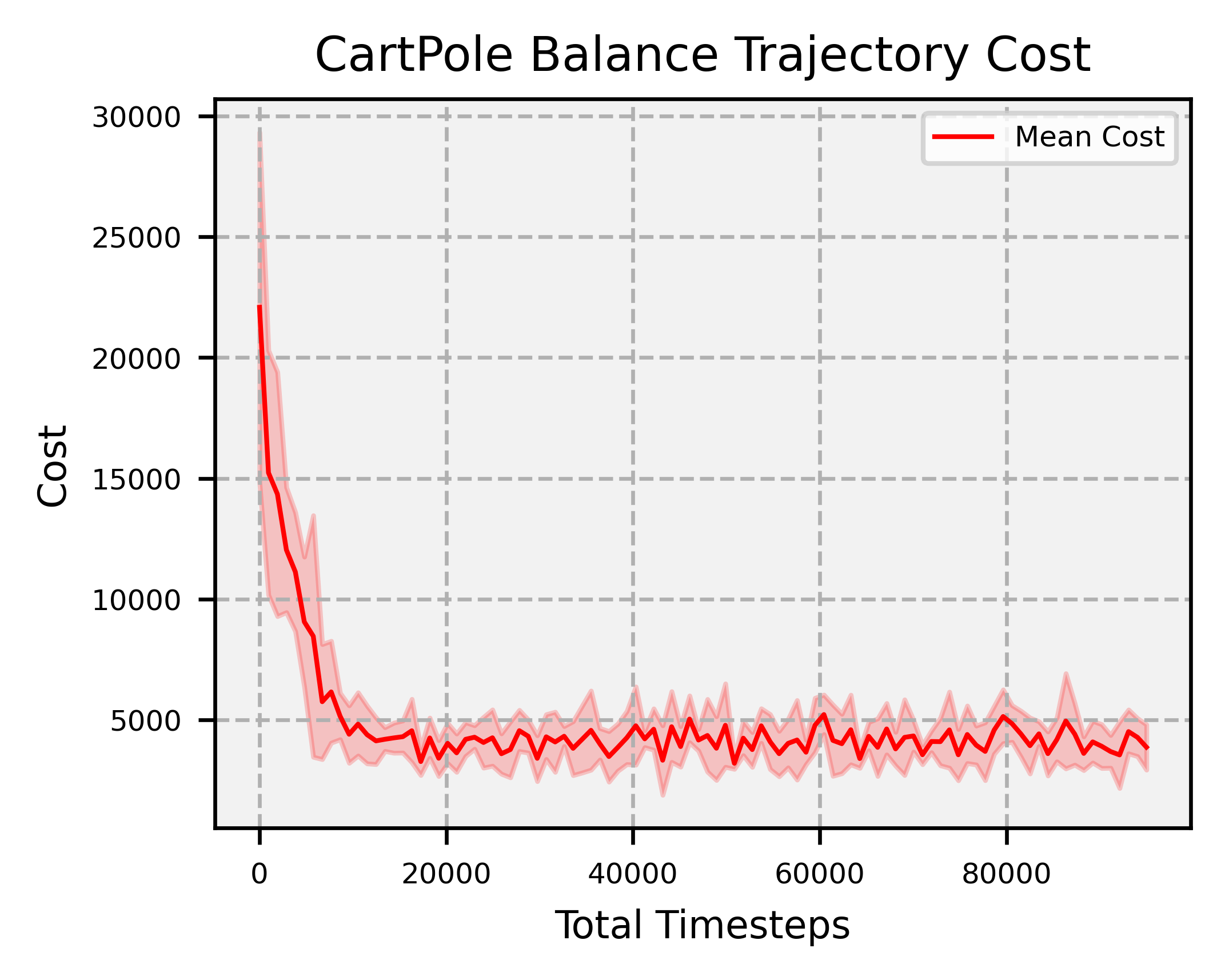}
  \end{minipage}\hfill
  \begin{minipage}{0.23\textwidth}
    \includegraphics[width=\textwidth]{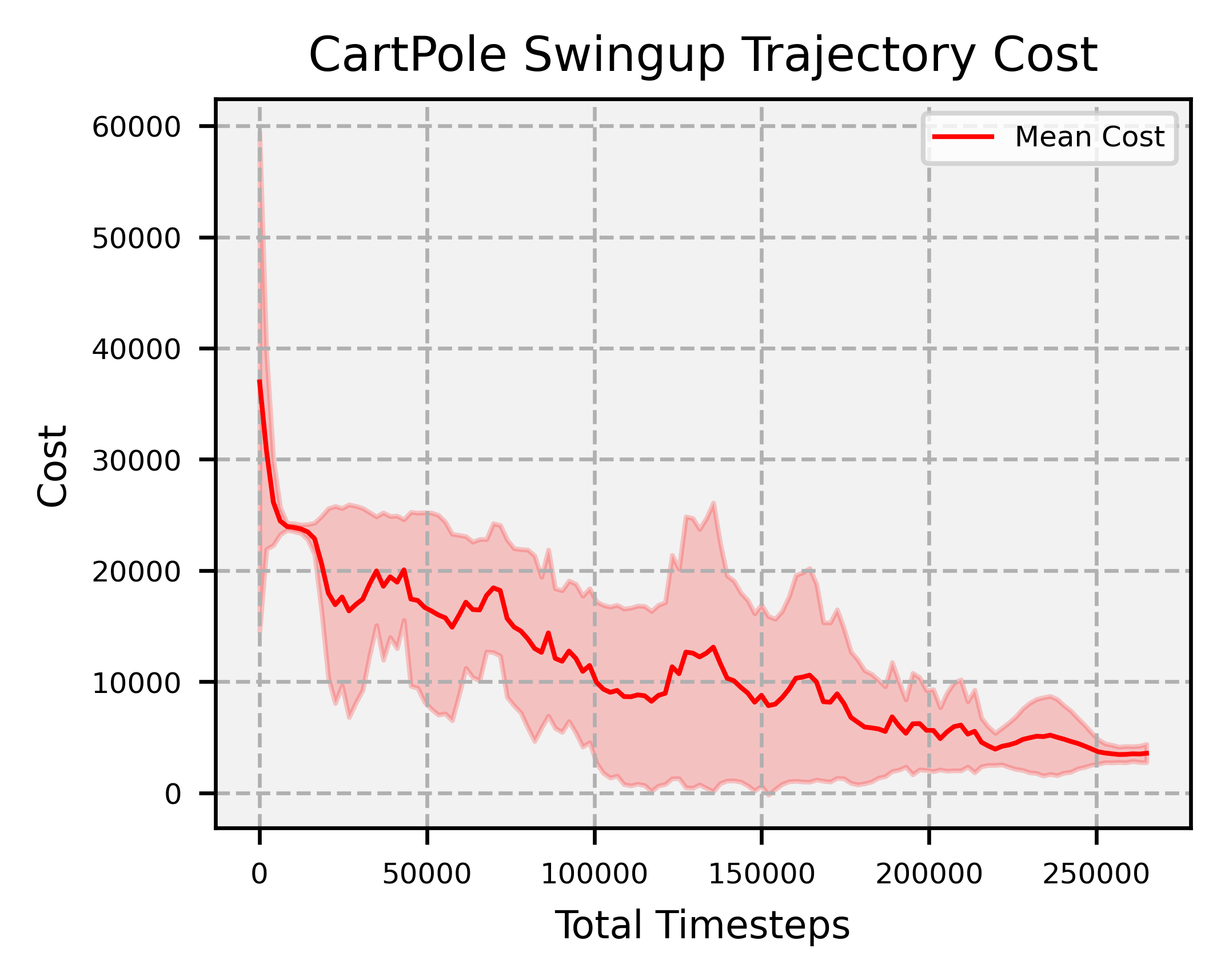}
  \end{minipage}\hfill
  \begin{minipage}{0.23\textwidth}
    \includegraphics[width=\textwidth]{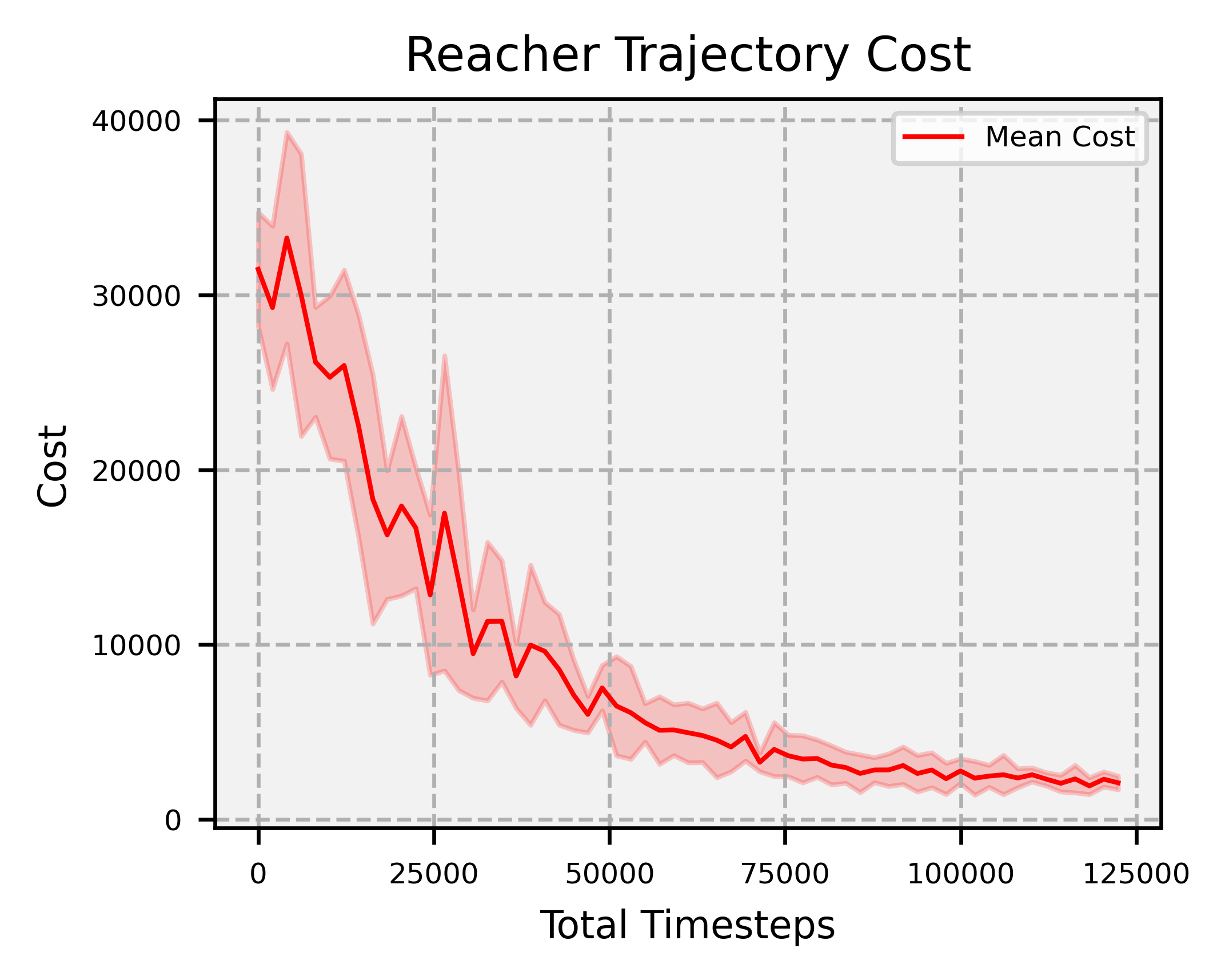}
  \end{minipage}

  \begin{minipage}{0.23\textwidth}
    \includegraphics[width=\textwidth]{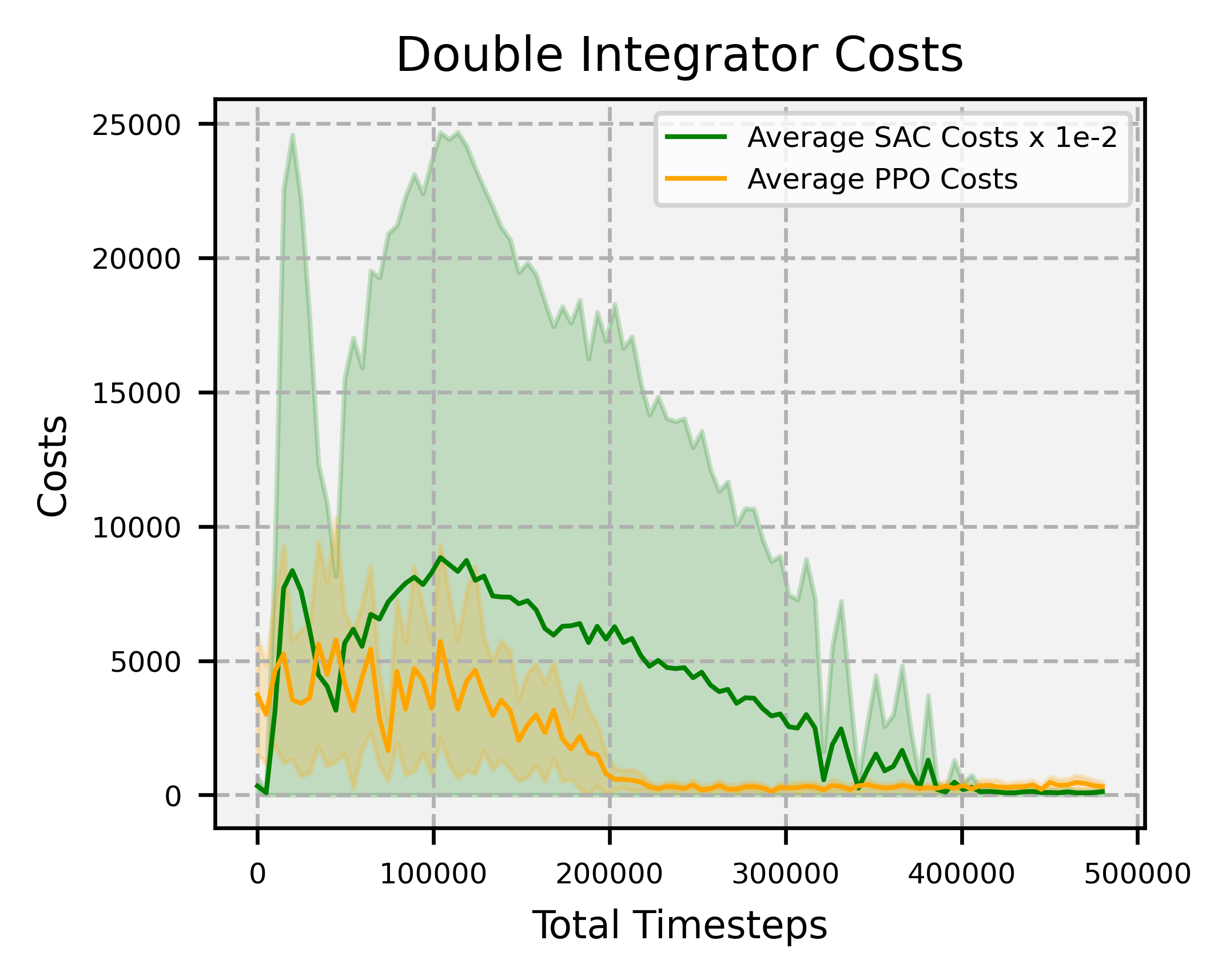}
  \end{minipage}\hfill
  \begin{minipage}{0.23\textwidth}
    \includegraphics[width=\textwidth]{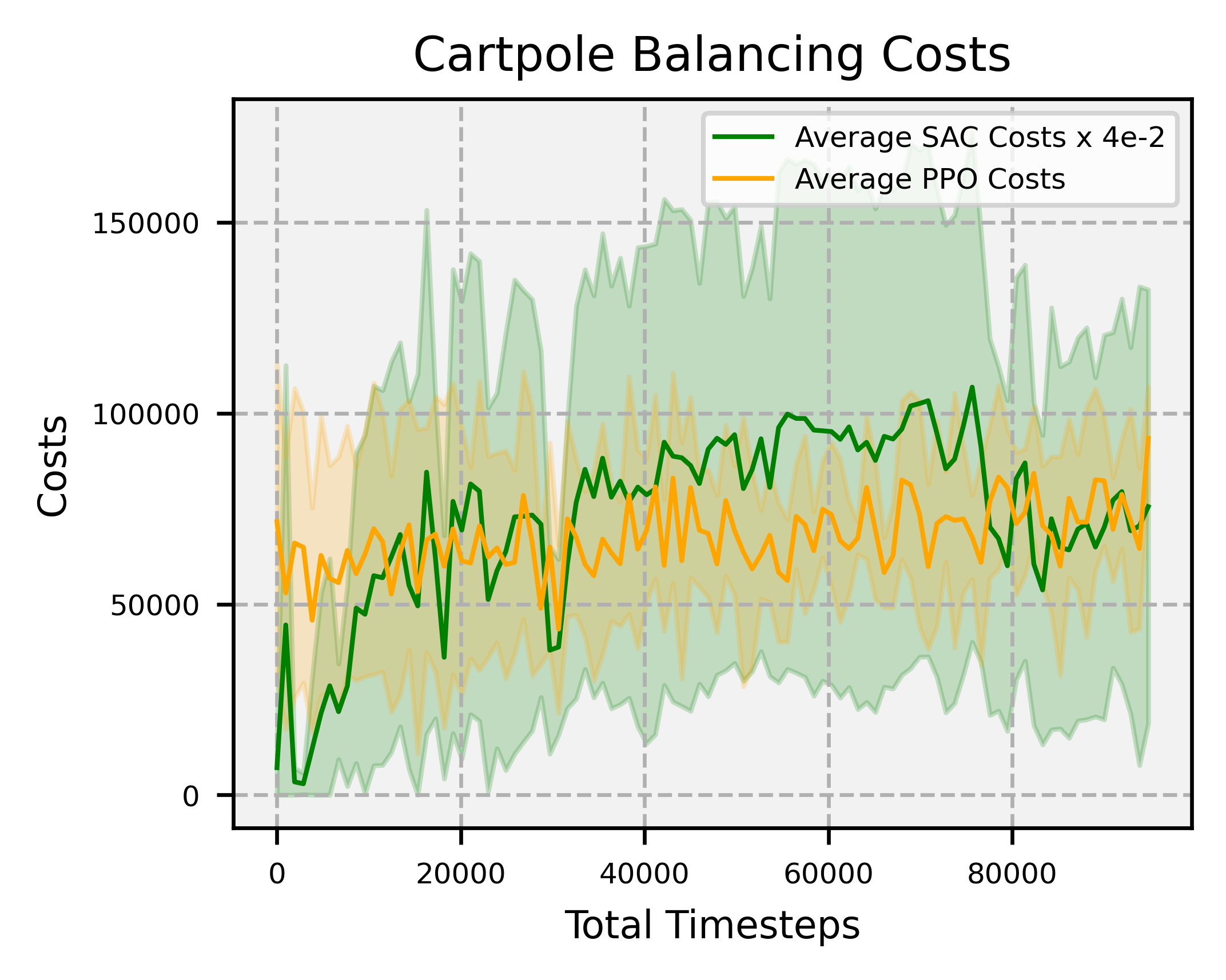}
  \end{minipage}\hfill
  \begin{minipage}{0.23\textwidth}
    \includegraphics[width=\textwidth]{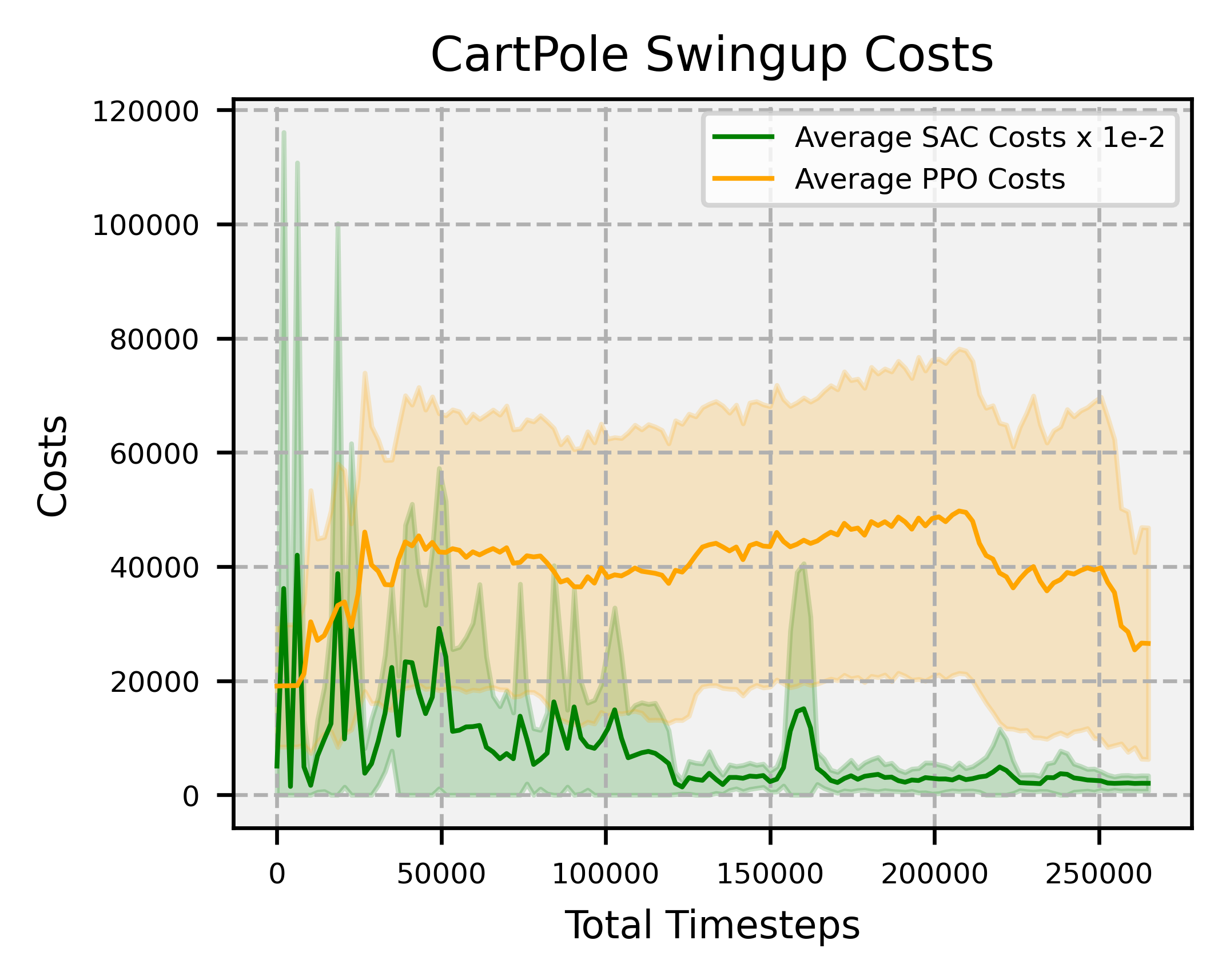}
  \end{minipage}\hfill
  \begin{minipage}{0.23\textwidth}
    \includegraphics[width=\textwidth]{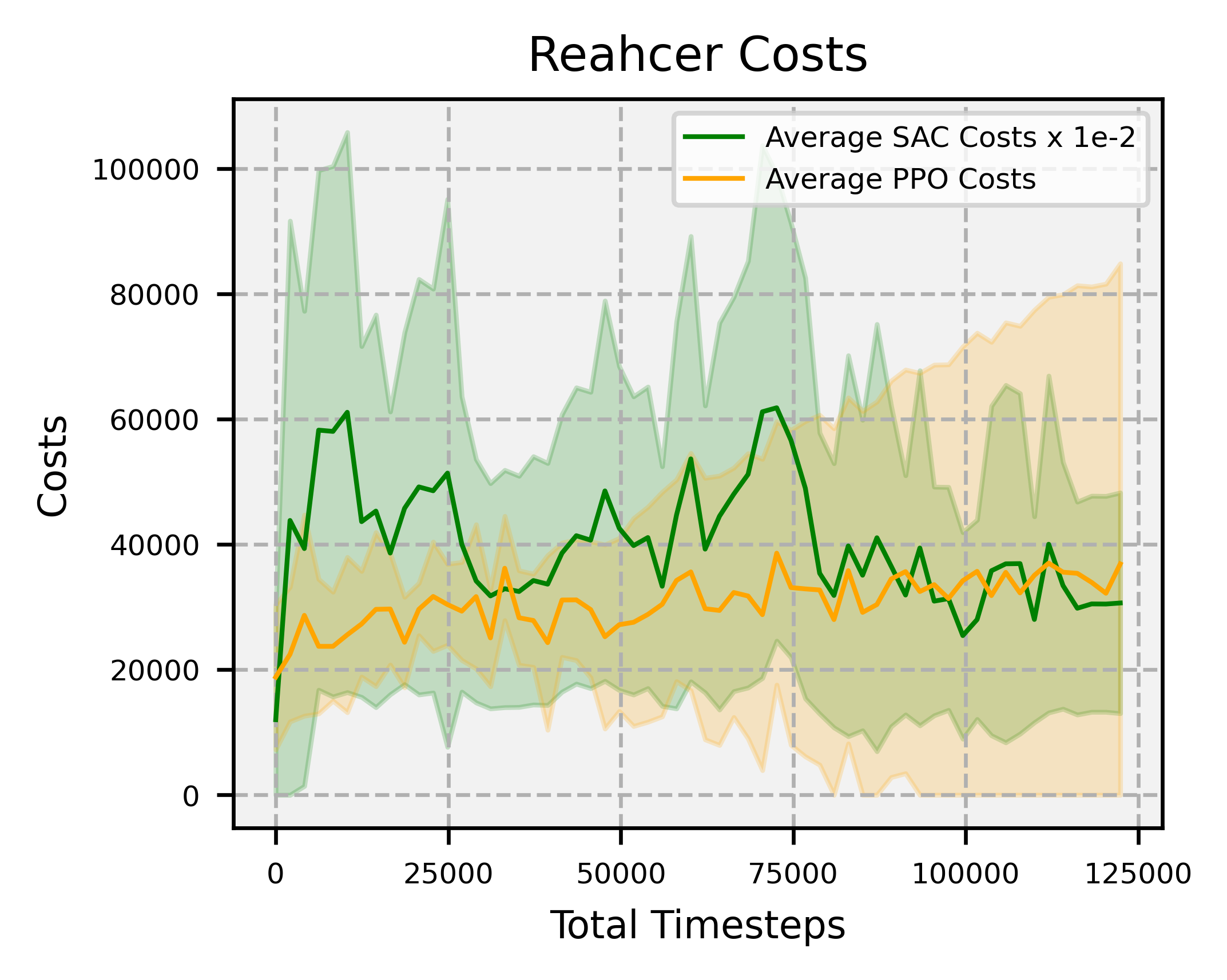}
  \end{minipage}
  \caption{Top row: Constraint satisfaction loss for value and Lyapunov function constraints. Middle row: Trajectory cost using our method. Bottom row: SAC and PPO trajectory cost. \textbf{Due to high values, SAC costs are scaled for visualisation}.}
  \label{fig:plots}
\end{figure*}

To evaluate our method, we apply it to tasks with linear or locally linear dynamics, such as the Double Integrator and Cartpole stabilisation. We also consider nonlinear problems like Cartpole swing-up and the planar reacher environment. Our method successfully learns time-varying Lyapunov and value functions that approximately satisfy the Bellman and Lyapunov constraints through minimising programs \ref{eq: value prog} and \ref{eq: lyap prog}. For a comprehensive assessment, we compare the effectiveness of the learned functions against Soft Actor Critic (SAC) \cite{haarnoja2018soft} and Proximal Policy Optimisation (PPO) \cite{schulman2017proximal} using the Stable-Baselines3 implementation \cite{stable-baselines3}. The hyperparameters for the baseline experiments can be found in our supplementary material. The results are presented in accompanying table \ref{table:performance_comparison} and figure \ref{fig:plots}.
For all tasks, we used our implementation of neural ODEs gradient estimator defined in equation \ref{eq: node pmp}. The results were obtained on a core i9 Intel processor with Nvidia GeForce RTX 4070 GPU. The Cartpole and the Reacher models are adopted from \cite{underactuated}\cite{1279748} 
\subsubsection{Environments and solver setting}\leavevmode\\
\textbf{Dynamics:}\label{Dynamics} Since our method leverages first-order gradients, we require gradients to be available within the PyTorch graph. As a result, we implement our own dynamics.

In the introduction we mentioned convergence strategies such as dampening dynamics and avoiding areas of the state and control space. Rigid body dynamics is defined by $\mathbf{M}(\vec q) \ddot{\vec q} + \mathbf{C}(\vec q, \dot{\vec q})\dot{\vec q} + \mathbf{b}(\dot{\vec q}) + \mathbf{G}(\vec q) = \mathbf{\vec \tau}$. However, in the case of very high friction: $\mathbf{C}(\vec q, \dot{\vec q})\dot{\vec q} \ll \mathbf{b}(\dot{\vec q})$, and the resulting dynamics becomes $\mathbf{M}(\vec q) \ddot{\vec q} + \mathbf{b}(\dot{\vec q}) + \mathbf{G}(\vec q)$. Thus by tuning the environment with high friction coefficients, one can practically remove a significant nonlinearity, as done in the Mujoco Reacher environment used in OpenAI's Gym \cite{1606.01540}.
With regards to the state space, for example in CartPole-V1, the control space is discrete where; 0 is push cart to the left and 1 is push cart to the right. Also, the termination condition is active not far from the upright where; $|\vec q_{pole}| \geq 0.2~\text{rads}$ and $|\vec q_{cart}| \geq 2.4$. For the Inverted Double Pendulum; the control space is constrained within $-1 \leq \vec u \geq 1$. The termination condition is active when the pole length falls below 1 and the upright height is 1.196; $l_1 \cos(\vec q_{pole1}) + l_2 \cos(\vec q_{pole1} + \vec q_{pole2}) \leq 1$. Similar termination conditions are also used on the continuous control CartPole-V4 environments. Although such strategies can help convergence, they lead to policies that are trained on small regions of the state space. Additionally, boundaries for such constraints are not clear apriori. For instance control constraints can directly impact the behaviour of the policy and in many cases we do not know the control bounds at which the optimal solution is achieved. Additonally, dampening dynamics can also linearise and simplify the problem. However, it can lead to policies that are trained on unrealistic dynamics. To evaluate the complete effectiveness of the methods we avoid such strategies in our experiments. In the next section we explain the details of our reward/cost design, control and space boundaries and termination conditions.
\\
\textbf{Solver setting:}
\label{Settings} 
We aim to keep the setting identical between solvers. As a result, timestep $\Delta t$, horizon/episode length $T$, number of timestep, and rewards/cost are equal between solvers. Additionally, we make the control space unbounded $\vec u \in \mathbb{R}$, however, we regularise this space with quadratic regularisation $\|\vec u\|_{\vec R}$, $\vec R$ is also identical between solvers per task. Our reward design is the negative of the cost function. Where our cost functions are the simple sum of running quadratic functions, $\text{cost} = \vec x^\top(t) \vec Q \vec x(t) + \vec x^\top(T) \vec Q_T \vec x(T)$. We also do not employ any state-based termination conditions and only rely on the horizon to terminate. We do however try to tune the hyperparameters of each baseline solver to the best of our ability, starting from the parameters mentioned in the original papers \cite{schulman2017proximal} \cite{haarnoja2018soft}.

\begin{table*}
    \centering
    \begin{tabular}{cccccccc}
        \toprule
        \multirow{2}{*}{Environment} & \multirow{2}{*}{Constraint Loss} & \multicolumn{3}{c}{Trajectory Cost} & \multicolumn{2}{c}{Cost Improvement} & \multirow{2}{*}{Horizon} \\
        \cmidrule(lr){3-5} \cmidrule(lr){6-7}
        & & Ours & PPO & SAC & PPO & SAC & \\
        \midrule
        Reacher &  \(33.61 \pm 20.25\) & \(2086.24 \pm 419.84\) & \(15514.49, \pm 4465.52\) & \(3831267.56 \pm 471973.23\) & 7.43 & 1836.65 & 170 \\
        Swing up & \(24.65 \pm 16.29\) & \(3568.74 \pm 887.16\) & \(33147.12 \pm 16041.53\) & \(255854.68, \pm 81662.83\) & 144.47 & 10379.47 & 171 \\
        Balancing & \(0.07 \pm 0.12\) & \(3868.80 \pm 1012.27\) & \(93109.16 \pm 15629.57\) & \(2360774.63, \pm 1098978.99\) & 24.07 & 284.12 & 79 \\
        Double Integrator & \(0.03 \pm 0.03\) & \(178.92 \pm 77.63\) & \(359.81, \pm 128.76\) & \(13311.58, \pm 7681.00\) & 2.01 & 74.78 & 400 \\
        \bottomrule
    \end{tabular}
    \caption{Training statistics and performance comparison against SAC and PPO}
    \label{table:performance_comparison}
\end{table*}
\subsection{Value Function Results}
\subsubsection{Reacher} For this fully actuated system, the goal state is defined as $\vec x = [0, 0, 0, 0]$. It is important to mention the dynamics here follows the low friction definition mentioned in Section \ref {Dynamics}. The cost function is quadratic with respect to $\vec{x} \in \R^4$ and control $\vec{u} \in \R^1$ with $\vec Q = \text{diag}(1, 1, 0, 0)$ and $\vec R = \vec M^{-1}(\vec q)$ where $\vec M^{-1}(\vec q)$ is the inverse inertial matrix. Additionally we use a terminal cost $\vec Q_T = \text{diag}(100, 100, 1, 1)$. The negative of this cost is used as a reward. The value function is fully connected neural network (FCN) $\Tilde{V}:\R^{4+1} \rightarrow \R^1$: (5-128-128-1 FCN). All discretisation timesteps are 0.01s and the Adam optimiser for training \cite{kingma2014adam}. As shown in table \ref{table:performance_comparison}, we satisfy the HJB constraint \ref{eq: dvdt} with an error of 0.25 $\pm 0.10$. Constraint satisfaction approximately converges at 5e4, however, small errors result in task cost converging at approximately 1e5 timesteps. Our average final cost outperforms SAC and PPO by a factor of 7.43 and 1836.65.
\subsubsection{Cartpole Swing Up} The Cartpole provides an underactuated environment obtained from \cite{underactuated}. In this, the goal is defined at the unstable equilibrium $\vec x = [0, 0, 0, 0]$. A the quadratic cost function is used with respect to $\vec{x} \in \R^4$ and control $\vec{u} \in \R^1$ with $\vec Q = \text{diag}(0, 0, 0, 0)$ and $\vec R = \vec M^{-1}(\vec q)$. Additionally, we use a terminal cost $\vec Q_T = \text{diag}(80, 600, .8, 4.5)$. The negative of this cost was used for baselines. The value function is parameterised by $\Tilde{V}_{tl}:\R^{4+1} \rightarrow \R^1$: (5-128-128-1 FCN). The program was solved using the same timestep and optimiser as Reacher. The program satisfied HJB constraint by 24.65. Constraint satisfaction approximately converges at 2e5 with small errors causing cost convergence at under 2.5e5 timesteps. As shown in table \ref{table:performance_comparison}
SAC and PPO show no convergence in this time.
\subsection{Lyapunov Function Results}
\subsubsection{Double Integrator}
We also assess the ability of program \ref{eq: lyap prog} to satisfy trajectory stability Lyapunov constraints on locally linear systems. We consider a fully linear double integrator with the goal state at $\vec x = [0, 0]$. For this problem we require the rate of change of the stable trajectories to be upper bounded by the quadratic cost with respect to $\vec{x} \in \R^2$ and control $\vec{u} \in \R^1$ with $\vec Q = \text{diag}(10, 0.1)$, $\vec Q_T = \text{diag}(10, 0.1)$ and $\vec R = \vec M^{-1}(\vec q)$ with the Lyapunov function $\Tilde{V}:\R^{2+1} \rightarrow \R^1$: (3-64-64-1 ICNN). Our results show that we satisfy the Lyapunov constraint \ref{eq: dvdt<} with error 0.03 $\pm$ 0.03. Additionally, we show that if we keep the set of initial conditions  $\vec X_0$ we are able guarantee Lyapunov stability of 90$\%$ of the compact set of K = $100$ initial conditions. This is also shown in the figure \ref{fig:DI Lyap}. Similar to the previous cases constraint satisfaction loss converges quicker than trajectory cost. Surprisingly SAC significantly underperforms on this task. We outperform SAC and PPO by a factor of 74.78 and 2.01.
\subsubsection{Cartpole Balancing}
We apply program \ref{eq: lyap prog} locally linear task, Cartpole balancing. Again we do not terminate episode based on any state \ref{Settings}. Our termination is only based on the end of Horizon. The goal state is at $\vec x = [0, 0]$ with initial conditions $\vec X_0 \in (-0.6, 0.6)$. Similarly, rate of change of the stable trajectories to be upper bounded by the quadratic cost with respect to $\vec{x} \in \R^4$ and control $\vec{u} \in \R^1$ with $\vec Q = \text{diag}(0, 25, 0.5, 0.1)$, $\vec Q_T = \text{diag}(0, 25, 0.5, 0.1)$ and $\vec R = \vec M(\vec q)^{-1}$ with the Lyapunov function $\Tilde{V}:\R^{4+1} \rightarrow \R^1$: (5-200-500-1 ICNN). In this, the Lyapunov constraint is satisfied \ref{eq: dvdt<} with error 0.17 $\pm$ 0.16.
Additionally, the same program can satisfy Lyapunov stability of 85$\%$ of the compact set of K = $100$ constant initial conditions. Constraint and task cost both converge at approximately 3e4 timesteps, with the final cost reaching 3868.80, outperforming SAC and PPO by a factor of 284.12 and 24.07.
\begin{figure}[h]
  \centering
  \includegraphics[width=1\linewidth]{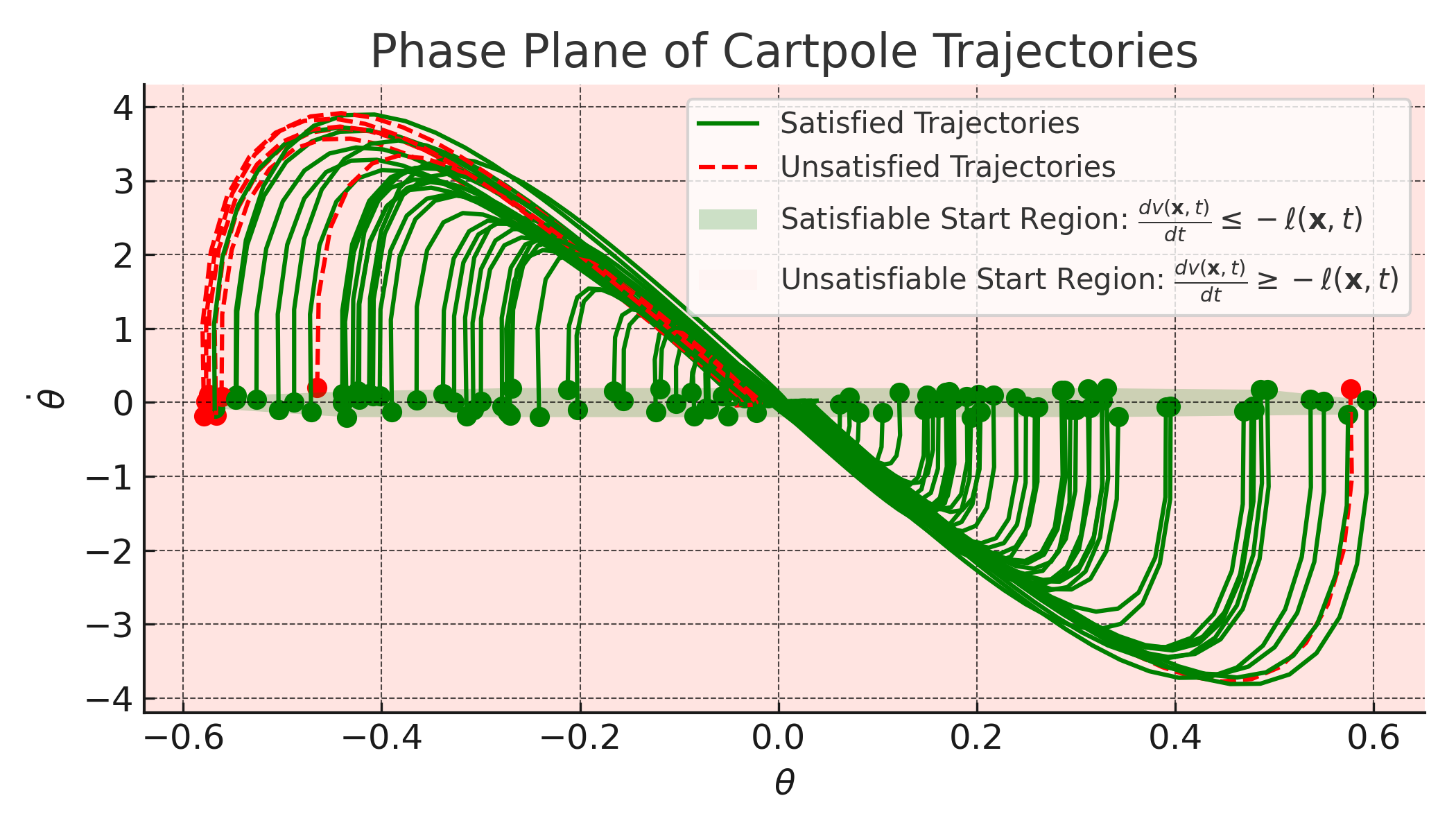}
  \caption{Cartpole balancing Lyapunov trajectories.}
  \label{fig:CP Lyap}
\end{figure}
\section{Discussion And Future Work}
\label{sec:disc}
The results show that the proposed programs notably outperform both SAC and PPO in terms of: convergence, costs associated with generated solutions, variance in results, and training stability. This difference is largely attributed to leveraging dynamics and its derivative information within training, analytically encoding dynamic structure within the policy, deterministic dynamics, and time parameterisation of the policy. In our evaluation, we treated the dynamics models as exact. This is a widespread practice within OC, because of the certainty equivalence assumption. This is where state feedback and re-optimisation through methods like Model Predictive Control can compensate for surprising amounts of error \cite{recht2019tour}. Derivative information through optimisation allows for considering the dynamic effects when minimising task cost. Additionally, the analytical inclusion of model parameters essentially compensates for the dynamic effects of the system, and reduces the complexity of the search space. Typically, RL algorithms are optimised over fixed horizons or episode lengths but do not consider this time dependency within the policy. This essentially results in using an infinite-horizon policy for a finite-time task, which is provably sub-optimal.

There are also challenges with the proposed programs. The HJB constraint and its relaxation play a central role in our proposed approach. The programs \ref{eq: value prog} and \ref{eq: lyap prog} essentially aim to learn the underlying governing function of the differential equation \ref{eq: dvdt} and its relaxation. Our results show that we can satisfy these constraints over a compact set, albeit up to a margin. The existence of this margin can have non-trivial impacts. For example, small approximation errors within the value function can have larger impacts on the task cost. We hypothesise that the task cost slower convergence is a result of this phenomenon. Additionally, this margin can pose challenges in safety-critical tasks where Lyapunov guarantees are required and exact state information is not available. The credit assignment problem, in our case, remains an issue. In our results, the HJB constraint is defined through the assigned cost function. This cost function essentially encodes the optimality constraint and, if defined poorly, can result in poor task performance.
The choice of time horizon is also non trivial. In our experiments we initially perform line search over a select number of horizons and select the best one. However, our method is robust to a variety of horizons so long as values or not chosen to be very high or low. Additionally, the discretisation time $\Delta t$ is also chosen by considering both simulation efficiency and approximation errors.

Finally, our method is online which can lead to typical problems associated with online algorithms such as non-stationary learning which can result in forgetting, especially in long horizon tasks.

We aim to extend this work in other further practical directions. For instance, the problem formulation and the environments used were all deterministic. This work can be extended to the stochastic domain where the benefits of sampling for example smoothing or exploration can be evaluated. Additionally, in this work, we primarily focused on problem formulation and empirical verification of the proposal. However, in further work, we aim to evaluate this method in more complex discontinuous environments with contact dynamics. However it is important to mention the ability to do this requires the availability of differentiable GPU based simulators which are currently in early stages of development. Finally, in our method, we derived the HJB policy under quadratic regularisation. This can be a limiting factor in cases where different input constraints are required. In future work, we aim to show that this extension can be relatively trivial as analytical policies are computable for any convex positive semi-definite function in $\vec u$ \cite{lutter2021value}.
\section{Conclusion}
We started by asking two questions. Can widely used RL algorithms such as SAC and PPO solve simple continuous control problems with minimal reward and dynamics shaping? Our results show that RL still requires significant effort in tuning to be able to achieve reasonable performance. Our second question asks for an alternative. We answer this by introducing two mathematical programs that use the HJB equation and first order gradients to learn value and Lyapunov functions. We demonstrate our effectiveness empirically by comparing to PPO and SAC on linear and nonlinear control-affine tasks. Our results show that we can outperform both in terms of quality of the generated solution, task cost, variance in results and training stability. 

\addtolength{\textheight}{-5.cm}   


\bibliographystyle{IEEEtran}
\bibliography{references}  
\end{document}